\documentclass{article}
\usepackage{spconf,amsmath,epsfig}
\usepackage{amsfonts}
\usepackage{algorithm, algorithmic}

\usepackage[title]{appendix}
\usepackage{color}
\usepackage{multirow}
\usepackage{amsmath}
\usepackage{breqn}
\usepackage{amssymb}
\usepackage{xcolor}
\usepackage{latexsym}

\let\OLDthebibliography\thebibliography
\renewcommand\thebibliography[1]{
  \OLDthebibliography{#1}
  \setlength{\parskip}{0pt}
  \setlength{\itemsep}{0pt plus 0.3ex}
}

\newcommand{\methodname}{{\tt{PAS-AFL}}}

\begin{document}\sloppy

\def\x{{\mathbf x}}
\def\L{{\cal L}}

\title{Agent-oriented Joint Decision Support for Data Owners \\ in Auction-based Federated Learning}
%

\name{Xiaoli Tang$^1$, Han Yu$^1$, Xiaoxiao Li$^2$}
\address{$^1$College of Computing and Data Science, Nanyang Technological University, Singapore\\
$^2$Department of Electrical and Computer Engineering, The University of British Columbia, Canada\\
\{xiaoli001, han.yu\}@ntu.edu.sg, xiaoxiao.li@ece.ubc.ca
}

\maketitle

\begin{abstract}
Auction-based Federated Learning (AFL) has attracted extensive research interest due to its ability to motivate data owners (DOs) to join FL through economic means. While many existing AFL methods focus on providing decision support to model tusers (MUs) and the AFL auctioneer, decision support for data owners remains open. To bridge this gap, we propose a first-of-its-kind agent-oriented joint \underline{P}ricing, \underline{A}cceptance and \underline{S}ub-delegation decision support approach for data owners in AFL (\methodname{}). 
By considering a DO’s current reputation, pending FL tasks, willingness to train FL models, and its trust relationships with other DOs, it provides a systematic approach for a DO to make joint decisions on AFL bid acceptance, task sub-delegation and pricing based on Lyapunov optimization to maximize its utility.
It is the first to enable each DO to take on multiple FL tasks simultaneously to earn higher income for DOs and enhance the throughput of FL tasks in the AFL ecosystem.
Extensive experiments based on six benchmarking datasets demonstrate significant advantages of \methodname{} compared to six alternative strategies, beating the best baseline by 28.77\% and 2.64\% on average in terms of utility and test accuracy of the resulting FL models, respectively.
\end{abstract}

\section{Introduction}
\label{sec:introduction}
In recent years, auction-based federated learning (AFL) has attracted significant research attention due to its ability to efficiently incentivise data owners (DOs) to participate in federated model training \cite{tang2023utility,tang2023competitive,zhang2021incentive}. In the context of AFL, the principal actors include the auctioneer, DOs, and model users (MUs). The auctioneer functions as an intermediary, disseminating bid requests from DOs to MUs, who subsequently submit their bids if they are interested. The auctioneer further consolidates the auction outcomes and communicates them back to the DOs. 
DOs take part in auctions by offering their data and compute resources to participate in the FL training process for the winning MUs. MUs submit bids to the auctioneer, receive the auction results, and orchestrate the FL training process with the recruited DOs.

Existing AFL methods can be divided into two categories based on their primary foci \cite{tang2023utility}: 1) MU-side methods, and 2) auctioneer-side methods.
The MU-side methods are generally designed to help MUs select and bid for DOs to optimize key performance indicators (KPIs) within budget constraints. 
The auctioneer-side methods are generally designed to optimize MU-DO pairing and pricing strategies, along with selecting suitable auction mechanisms. The overarching goal is to achieve specific system-level objectives, such as maximizing social welfare or minimizing social costs. 
The challenges on the DO-side remain open. This lack of decision support for DOs, regarding resource allocation and reserve price setting, can lead to sub-optimal FL participation decisions. 

To bridge this important gap, we propose the agent-oriented joint \underline{P}ricing, \underline{A}cceptance and \underline{S}ub-delegation decision support approach for data owners in AFL (\methodname{}) to provide decision support for DOs in AFL ecosystems. 
Focused on individual-level decision-making for DOs, it is designed based on Lyapunov optimization \cite{yu2016mitigating,yu2017algorithmic} with the aim to achieve superlinear time-averaged collective activeness within the AFL ecosystem.
\methodname{} takes a DO's current reputation, pending FL tasks, willingness to engage in FL model training, and trust relationships with other DOs into account. It offers a systematic framework for DOs to make informed decisions regarding FL training task acceptance, sub-delegation, and pricing. The primary goal is to maximize utility, while mitigating fluctuations in the delay of pending FL tasks. In addition, it provides an interpretable control variable that allows DOs to fine-tune the balance between utility improvement and the costs incurred by delay mitigation actions.

To the best of our knowledge, \methodname{} is the first decision support method proposed for AFL DOs. It can be deployed as a personal assistant agent for each DO. Extensive experiments based on six benchmarking datasets demonstrate significant advantages of \methodname{} compared to six alternative strategies, beating the best baseline by 28.77\% and 2.64\% on average in terms of utility and test accuracy of the resulting FL models, respectively.


\section{Related Work}
\label{sec:related_work}

Methods designed for the entire AFL ecosystem generally aim to maximize social welfare, minimize social cost and maximize the total utility of all MUs or DOs \cite{li2019credit, krishnaraj2022future,hong2020optimizing,gao2019auction, yang2020task}. These methods often adopt double auction or combinatorial auction to determine the optimal DO-MU matching and pricing. 
Methods for one single AFL MU can be further refined into two subcategories: i) reverse auction-based methods, and ii) forward auction-based methods.
Reverse auction-based methods \cite{jiao2020toward, zeng2020fmore, ying2020double, zhang2021incentive,deng2021fair} tackle the challenge of DO selection within a reverse auction setting with the aim of achieving the key performance indicators of the MU. Forward auction-based methods \cite{tang2023utility,tang2023competitive} focus on helping MUs bid for DOs from the same pool.  
Different from existing methods, \methodname{} is designed to serve the goals of a DO by helping it determine the number of FL tasks to accept, to sub-delegate, and how to price the resources it offers.

\section{Preliminary}
\label{sec:preliminary}
Let $\mathcal{N}$ represent the set of DOs within an AFL ecosystem. We study them over a series of time steps $\mathcal{T} = \{0, 1, 2, \cdots, t, \cdots, T-1\}$. 
In each time step $t \in \mathcal{T}$, DO $i \in \mathcal{N}$ can accept $\kappa_i(t) < \hat{\kappa}_i^{max}$ FL tasks from MUs via the auction process. $\hat{\kappa}_i^{max}$ denotes the maximum number of tasks $i$ can accept during the auction process. The unit payment for each task, transferring from MUs to $i$, is denoted as $p_i(t)$. Once $i$ accepts a training task, it must either complete it within a specified time frame, or sub-delegate it to other DOs (e.g., to avoid penalties for missing the deadline and potential damage to its reputation \cite{shin2022fedbalancer}). 

\begin{figure}[t!]
\centering
\includegraphics[width=0.8\columnwidth]{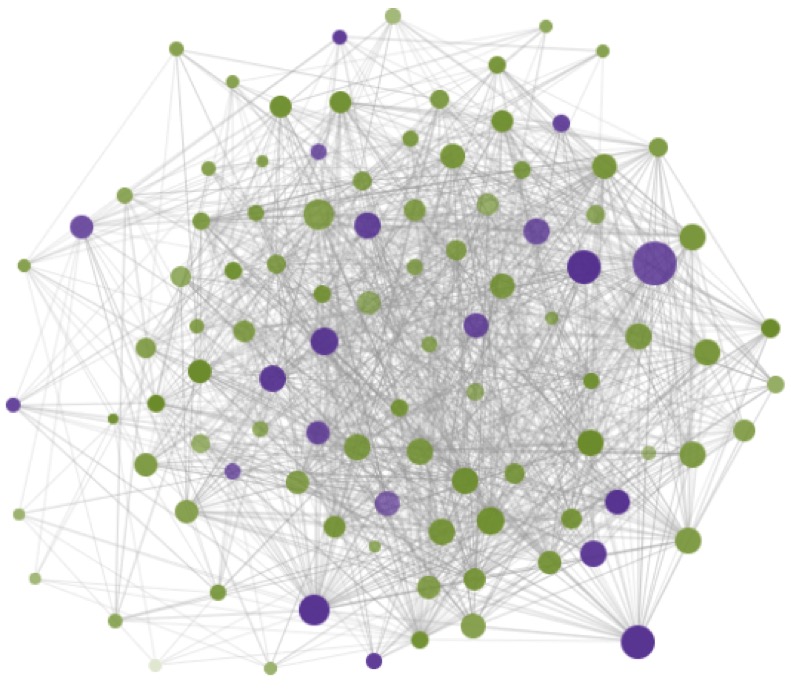}
\caption{An example trust network among agents. The purple nodes denote the agents who are delegating/sub-delegating FL tasks. The green nodes represent the DO agents. The size of the node corresponds to its connectivity \cite{gao2023multi}.}
\label{fig:node_network}
\end{figure}

Reputation $r_i(t)$ for $i$ is important for establishing trust-based connections ($\boldsymbol{n}_i$) with other DOs \cite{gao2023multi}, as shown in Figure \ref{fig:node_network}. $\boldsymbol{n}_i$ consists of trusted DOs to whom $i$'s FL tasks can be delegate to. It also serves as a measure of how reliable $i$ is, which affects its future prospect of being selected. The expected future demand on $i$'s local data, $\mathbb{E}(\kappa_i(t))$, can be formulated as:
\begin{equation}
\small
\label{eq:future_demand}
\begin{aligned}
\mathbb{E}(\kappa_i(t))= f(r_i(t), p_i(t)).
\end{aligned}
\end{equation}
According to empirical studies in the field of e-commerce \cite{ye2013depth}, $f(r_i(t), p_i(t))$ satisfies the following relationship:
\begin{equation}
\small
\label{eq:demand_for_service}
\begin{aligned}
\ln f(r_i(t), p_i(t)) =  & a_o - a_1 \ln r_i(t) + a_2 \ln M^p_i(t) + a_3 \epsilon_i  \\
& \quad + \ln p_i(t), 
\end{aligned}
\end{equation}
where the constants $a_0$, $a_1$, $a_2$, and $a_3$ are positive values. $\epsilon_i$ quantifies the degree of alignment between the quality of training models delivered by data owner $i$ and the quality it promises. $M^p_i(t)$ denotes the number of positive ratings that data owner $i$ has received within a specified time interval. Eq. \eqref{eq:demand_for_service} can be re-expressed as:
\begin{equation}
\small
\label{eq:demand_for_service_1}
\begin{aligned}
f(r_i(t), p_i(t)) =  \zeta_i \frac{p_i(t)}{(r_i(t))^{a_1}},  \quad \zeta_i = e^{a_0 +a_3 \epsilon_i}(M^p_i(t))^{a_2}.
\end{aligned}
\end{equation}
 
 
\section{The Proposed \methodname{} Approach}
\label{sec:model}
\begin{figure}[t!]
\centering
\includegraphics[width=0.8\columnwidth]{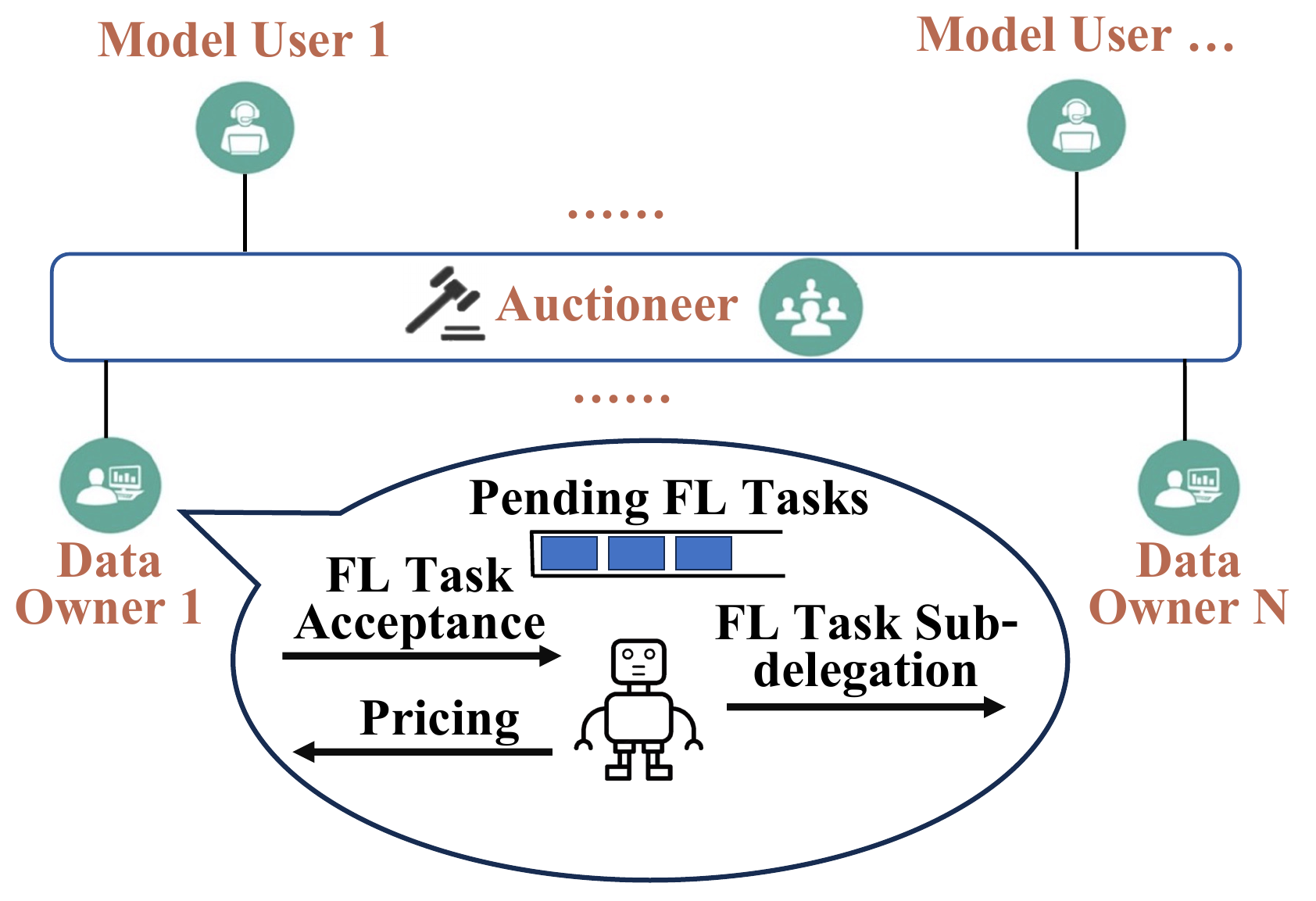}
\caption{The \methodname{} system architecture.}
\label{fig:model}
\end{figure}

The decision support offered by \methodname{} revolves around three key actions: 1) FL task acceptance, 2) FL task delegation and sub-delegation, and 3) pricing strategies (Figure \ref{fig:model}). It provides a systematic approach that empowers a DO $i$ to make joint decisions, taking into account factors including its current reputation, pending workload, willingness to engage in model training and its trust relationships with other DOs. 
\subsection{Modeling Data Owner Decision Dynamics}
Post auction, DO $i$ can proceed with the FL tasks according to its own schedule. To facilitate the management of pending FL tasks for $i$, we model such tasks by a virtual queue $q_i(t)$. 
The queuing dynamics for $q_i(t)$ is:
\begin{equation}
\small
\label{eq:virtual_pending_task_queue}
\begin{aligned}
q_i(t+1) = [q_i(t) - \theta_i(t) - s_i(t)]^+ + x_i(t)\kappa_i(t),
\end{aligned}
\end{equation}
where $[\cdot]^+ = \max[\cdot,0]$, and $\theta_i(t)$ represents the progress made by $i$ at time step $t$ on its pending FL tasks. It is worth noting that, due to the limit on the DO's processing capacity, $\theta_i(t) \leq \theta_i^{\max}$, where $\theta_i^{\max}$ is the maximum number of tasks that $i$ can process at each time step. Similarly, $s_i(t) \leq s_i^{\max}$ denotes the number of tasks sub-delegated to other DOs by $i$ at time step $t$, with $s_i^{\max}$ representing the maximum allowable task sub-delegated by $i$ at each time step. As indicated by \cite{shi2023fedwm}, DOs may need to balance other essential tasks (e.g., collecting and labeling fresh data) with FL tasks. Therefore, as shown in Eq. \eqref{eq:virtual_pending_task_queue}, we introduce a binary decision variable $x_i(t)$ into \methodname{} to determine whether $i$ shall accept new FL tasks at the start of the time step $t$. 

To meet the timeline constraint of FL tasks and avoid potential penalties for failing to complete tasks before the deadlines, we introduce a conceptual queue $Q_i(t)$ to quantify the urgency for $i$ to sub-delegate its pending FL tasks. The dynamics of $Q_i(t)$ are:
\begin{equation}
\small
\label{eq:virtual_deadline_queue}
\begin{aligned}
Q_i(t+1) = [Q_i(t) - \theta_i(t) - s_i(t) + \bar{\kappa}_i \mathbb{I}_{(q_i(t) > 0)}]^+ ,
\end{aligned}
\end{equation}
where $\mathbb{I}_{(\text{condition})}$ is an indicator function. $\mathbb{I}_{(\text{condition})}$ equals to 1 if and only if the condition is true; otherwise, it equals to 0. $\bar{\kappa}_i$ is a constant to define the delay constraint for FL tasks. It is calculated as the average number of new FL tasks accepted by $i$ per time step:
\begin{equation}
\small
\label{eq:average_demand}
\begin{aligned}
\bar{\kappa}_i = \frac{1}{|T|}\sum_{t \in [0, T-1]} \kappa_i (t).
\end{aligned}
\end{equation}
The intuition behind the dynamics of $Q_i(t)$ in Eq. \eqref{eq:virtual_deadline_queue} is:
\begin{itemize}
    \item The pending tasks in $Q_i(t)$ is increased in such a manner that, if $q_i(t)$ is not empty at the moment when $Q_i(t)$ is updated, $Q_i(t)$ grows by $\bar{\kappa}_i$. This mechanism ensures that $Q_i(t)$ continues to rise if there are FL tasks in $q_i(t)$ that have remained incomplete over time.
    \item $Q_i(t)$ is decreased by FL task training and sub-delegation processes, denoted by $[\theta_i(t) + s_i(t)]$.
\end{itemize}

To maintain the sustainable operation of the auction ecosystem, ensuring stability in mean rates of virtual queues $Q_i(t)$ and $q_i(t)$ is crucial. We redefine the challenging time-coupling constraint as a new objective, setting an upper limit on queue growth to prevent indefinite expansion. The Lyapunov optimization technique \cite{yu2016mitigating} is leveraged for this purpose. Typically, a quadratic sum of the backlog in each queue serves as the Lyapunov function:
\begin{equation}
\small
\label{eq:lyapunov_quadratic}
\begin{aligned}
\mathcal{L}(\Theta_i(t)) = \frac{1}{2} [q^2_i(t) + Q^2_i(t)]. 
\end{aligned}
\end{equation}

To measure the expected increase of the Lyapunov function $\mathcal{L}(\Theta_i(t))$, we design a Lyapunov drift as:
\begin{equation}
\small
\label{eq:lyapunov_drift}
\begin{aligned}
\Delta(\Theta_i(t)) = \mathbb{E}[\mathcal{L}(\Theta_i(t+1)) - \mathcal{L}(\Theta_i(t))|\Theta_i(t)]. 
\end{aligned}
\end{equation}
It describes the expected change in the function over one-time slot based on the current state, where the expectation is based on queue evolution statistics. The Lyapunov drift theorem asserts that by minimizing $\Delta(\Theta_i(t))$ at each time slot $t$, all queues shall be stabilized. In addition, fair participation criteria are automatically satisfied when all virtual participation queues are stabilized.

Considering $(\max[a-b +c, 0])^2 \leq a^2+b^2+c^2+2a(c-b)$, $\forall a, b, c \geq 0$, we obtain: 
\begin{equation}
\small
\label{eq:one_drift}
\begin{aligned}
& \Delta(\Theta_i(t)) = \mathbb{E}[\mathcal{L}(\Theta_i(t+1)) - \mathcal{L}(\Theta_i(t))] \\
& = \mathbb{E}[\frac{1}{2}   [\left ([q_i(t) - \theta_i(t) - s_i(t)]^+ + \kappa_i(t)\right ) ^2 \\
& \quad + \left ([s_i(t) - \theta_i(t) - s_i(t) + \bar{\kappa}_i \mathbb{I}(q_i(t) > 0)]^+\right ) ^2 \\
& \quad - q_i(t)^2 - Q_i(t)^2 )]] \\
& \leq \mathbb{E}[\xi_i + q_i(t)(\kappa_i(t) - \theta_i(t) - s_i(t)) \\
& \quad + Q_i(t)(\bar{\kappa}_i - \theta_i(t) - s_i(t))]
\end{aligned}
\end{equation}
where $\xi_i =(\theta_i^{\max} + s_i^{\max})^2 + (\kappa_i^{\max})^2 $. 

\subsection{Optimal Decision Strategy} 
As described above, DO $i$ may need to sub-delegate some of its pending FL tasks to other trusted DOs $\boldsymbol{n}_i$ to meet the time constraint of these tasks. Let $\bar{p}_{\boldsymbol{n}_i}(t) = \frac{1}{|\boldsymbol{n}_i|} \sum_{k \in \boldsymbol{n}_i} p_k(t)$ represent the average price charged by $i$'s trusted DOs for training an FL task at time step $t$ (e.g., based on prior experience). The cost of FL task sub-delegation can be formulated as:
\begin{equation}
\small
\label{eq:subdelegation_cost}
\begin{aligned}
s^{cost} = \mathbb{E}[\bar{p}_{\boldsymbol{n}_i}(t)s_i(t)|\Theta_i(t)].
\end{aligned}
\end{equation}

Let $c_i(t)$ denote the cost incurred by $i$ for each unit of FL task, which can be calculated by $i$ in advance based on its historical FL tasks. Then, the cost of completing $\theta_i(t)$ FL tasks can be formulated as:
\begin{equation}
\small
\label{eq:training_cost}
\begin{aligned}
t^{cost} = c_i(t) \times \theta_i(t).
\end{aligned}
\end{equation}
Then, we can formulate the utility, $u_i(t)$, obtained by $i$ at time step $t$ from the auction market as follows:
\begin{equation}
\small
\label{eq:utility}
\begin{aligned}
& u_i(t) = x_i(t)p_i(t)r_i(t)f(r_i(t), p_i(t)) - t^{cost} - s^{cost}\\
 & = x_i(t)p_i(t)r_i(t)f(r_i(t), p_i(t)) - \bar{p}_{\boldsymbol{n}_i}(t)s_i(t)  - c_i(t)\theta_i(t).
\end{aligned}
\end{equation}

The goal of $i$ is to maximize its utility by taking part in AFL, minimizing significant changes in its pending FL tasks while meeting their deadlines. Therefore, we adopt a \textit{\{utility-drift\}} formulation to model the joint objective:
\begin{equation}
\small
\label{eq:goal}
\begin{aligned}
& \rho_i(t) u_i(t) - \Delta(\Theta_i(t)) \\
&  \leq \rho_i(t) \mathbb{E}[x_i(t)p_i(t)r_i(t)f(r_i(t), p_i(t)) -  \bar{p}_{\boldsymbol{n}_i}(t)s_i(t) \\
& - c_i(t)\theta_i(t)|\Theta_i(t)] -\xi_i -  Q_i(t) \mathbb{E}[\bar{\kappa}_i - \theta_i(t) - s_i(t)|\Theta_i(t)] \\
& - q_i(t)\mathbb{E}[x_i(t)f(r_i(t), p_i(t)) - \theta_i(t) - s_i(t)|\Theta_i(t)].
\end{aligned}
\end{equation}
As indicated in \cite{shi2023fedwm}, a DO may have different levels of availability during different times for training FL tasks. In this sense, we adopt $\rho_i(t)$  in Eq. \eqref{eq:goal} to reflect the general eagerness of DO $i$ to accept new FL tasks. A higher value of $\rho_i(t)$ means that data owner $i$ is more highly available to take on new FL tasks. In addition, as shown in Eq. \eqref{eq:goal}, $\rho_i(t)$ effectively modulates the relative importance attributed to the two components within the \textit{\{utility-drift\}} expression. This value can be deduced by monitoring $i$'s progress in FL training over a specific time frame or explicitly specified by $i$ to control \methodname{} \cite{shi2023fedwm}. 

At the start of time step $t$, \methodname{} scrutinizes $Q_i(t)$ and $q_i(t)$, in conjunction with $i$'s prevailing context tuple $\langle c_i(t), \kappa_i(t), \bar{p}_{\boldsymbol{n}_i}(t)\rangle$ in order to ascertain the optimal value of the number of FL tasks to be sub-delegated $s_i(t)$, pricing of its local data and compute resources $p_i(t)$, and the FL task acceptance decision $x_i(t)$ by solving the \textit{\{utility-drift\}} maximization problem in Eq. \eqref{eq:goal}. We focus exclusively on $s_i(t)$, $x_i(t)$ and $p_i(t)$, and reformulate Eq. \eqref{eq:goal} as:
\begin{equation}
\small
\label{eq:goal_reformulate}
\begin{aligned}
& \rho_i(t) u_i(t) - \Delta(\Theta_i(t)) 
\leq -s_i(t)[\rho_i(t)\bar{p}_{\boldsymbol{n}_i}(t) - q_i(t) - Q_i(t)]\\
&  -\xi_i - \theta_i(t)[\rho_i(t)c_i(t) - q_i(t) - Q_i(t)]\\
& + [\rho_i(t)x_i(t)p_i(t)r_i(t)f(r_i(t), p_i(t)) - q_i(t) x_i(t) f(r_i(t), p_i(t))]. 
\end{aligned}
\end{equation}

\textbf{FL Task Sub-delegation Decision}.
By focusing only on terms involving $s_i(t)$, we re-express Eq. \eqref{eq:goal} as:
\begin{equation}
\small
\label{eq:goal_1_final}
\begin{aligned}
& \min \frac{1}{T} \sum_{t \in [0, T-1],\sum_{i \in \mathcal{N}}} s_i(t)[\rho_i(t)\bar{p}_{\boldsymbol{n}_i}(t) - q_i(t) - Q_i(t)],\\
& s.t. \quad \quad \quad \quad 0 \leq s_i(t) \leq q_i(t), \\ 
& \quad \quad \exists k \in \boldsymbol{n}_i, \exists \tau_j \in q_i(t), r_k(t) \geq r_{\min}(t) \land p_k(t) \leq p_{\tau_j}.
\end{aligned}
\end{equation}
Here, $r_{\min}(t) \in [0, 1]$ is the reputation threshold set by $i$. $p_{\tau_j}$ is the payment $i$ received for the FL task $\tau_j$. 
The first constraint defined in Eq. \eqref{eq:goal_1_final} quantifies the requirement that the number of FL tasks sub-delegated must not exceed the total number of pending tasks held by $i$. The second constraint, on the other hand, serves to quantify the availability of qualified DOs within the trust network for FL task sub-delegation, as well as the economic benefit to $i$ when sub-delegating tasks in line with market prices.

From Eq. \eqref{eq:goal_1_final}, the optimal value of $s_i(t)$ is:
\begin{equation}
\small
\label{eq:optimal_s}
\begin{aligned}
s_i(t) = 
\begin{cases}
0,  & \text{if $\rho_i(t)\bar{p}_{\boldsymbol{n}_i}(t) -  q_i(t) - Q_i(t) \geq 0$}, \\
q_i(t) - \theta_i(t), & \text{otherwise}.
\end{cases}
\end{aligned}
\end{equation}
Eq. \eqref{eq:optimal_s} implies that when data owner $i$ exhibits a strong willingness to engage in model training, encounters a high cost for sub-delegation, possesses a low current pending workload, and the tasks in the pending queue have not been delayed for an extended period, $i$ should refrain from sub-delegating any FL tasks. Otherwise, $i$ should aim to sub-delegate FL tasks. However, the precise value of $s_i(t)$ is also contingent upon the satisfaction of the constraint specified in Eq. \eqref{eq:goal_1_final}. This constraint stipulates that there must be at least one DO $k$ within $i$'s trust network, denoted as $k \in \boldsymbol{n}_i$, who not only has a reputation exceeding the threshold value but also charges a price no more than what $i$ can offer for the FL task. This condition ensures that $i$ does not need to incur any financial loss when sub-delegating the task to $k$.


\begin{table*}[!t]
\centering
\caption{Performance comparison. The best and the second-best results are highlighted in \textbf{Bold} and \underline{underline}, respectively.} 
\resizebox*{0.84\linewidth}{!}{
\begin{tabular}{|*{13}{c|}}
\hline
\multirow{2}*{Method} & \multicolumn{2}{c|}{MNIST} & \multicolumn{2}{c|}{CIFAR-10} &\multicolumn{2}{c|}{FMNIST}& \multicolumn{2}{c|}{EMNISTD} & \multicolumn{2}{c|}{EMNISTL} & \multicolumn{2}{c|}{KMNIST} \\\cline{2-13}
{} & Utility &  Acc & Utility &  Acc & Utility &  Acc & Utility &  Acc & Utility &  Acc & Utility &  Acc \\\hline
Rand-Rand & 16.86 & 82.70 & 16.41 & 41.09 & 17.53 & 74.37 & 18.43 & 83.08 & 16.58 & 73.05 & 18.61 & 72.45\\
Rand-Greedy & 15.80 & 82.22 & 15.95 & 40.93 & 16.33 & 74.75 & 15.91 & 81.80 & 16.27 & 72.76 & 15.80 & 71.94\\
Ampp-Rand & 14.46 & 81.38 & 13.26 & 40.62 & 15.64 & 73.82 & 13.69 & 82.21 & 14.09 & 73.64 & 14.48 & 71.96\\
Ampp-Greedy & 13.34 & 80.57 & 13.12 & 39.39 & 14.12 & 73.34 & 13.40 & 81.28 & 13.20 & 71.40 & 13.57 & 71.87\\
Lin-Rand & \underline{17.25} & \underline{83.83} & \underline{18.45} & \underline{41.68} & \underline{18.24} & 74.49 & \underline{19.06} & \underline{84.32} & \underline{17.33} & 73.15 & \underline{18.93} & \underline{72.38}\\
Lin-Greedy & 16.89 & 83.52 & 17.92 & 41.53 & 16.77 & \underline{75.37} & 18.57 & 83.92 & 16.78 & \underline{74.61} & 18.58 & 72.04\\\hline
\methodname{} & \textbf{23.67} & \textbf{87.79} & \textbf{23.20} & \textbf{42.94} & \textbf{23.49} & \textbf{77.35} & \textbf{23.18} & \textbf{86.31} & \textbf{23.08} & \textbf{74.62} & \textbf{23.86} & \textbf{74.70}\\
w/o pricing (R) & 22.45 & 85.81 & 22.37 & 42.42 & 22.96 & 76.53 & 22.23 & 85.71 & 22.54 & 74.33 & 22.63 & 73.20\\
w/o pricing (A) & 21.97 & 85.08 & 21.56 & 41.82 & 22.84 & 76.12 & 21.49 & 85.28 & 21.69 & 74.29 & 22.08 & 72.29\\
w/o pricing  (L) & 23.09 & 85.97 & 22.83 & 42.53 & 23.11 & 76.84 & 22.77 & 85.82 & 22.89 & 74.49 & 23.48 & 73.96\\
w/o sub-delegation (R) & 22.59 & 86.51 & 22.87 & 42.11 & 23.29 & 76.45 & 22.64 & 86.02 & 22.75 & 74.56 & 23.31 & 73.34\\
w/o sub-delegation (G) & 21.84 & 85.25 & 22.32 & 41.75 & 22.69 & 75.86 & 22.01 & 85.19 & 22.24 & 74.31 & 21.53 & 72.37\\\hline
\end{tabular}
}
\label{tab:performance}
\end{table*}

\textbf{Pricing and FL Task Acceptance Decisions}.
Since \methodname{} primarily focuses on controlling the decision variables $p_i(t)$ and $x_i(t)$ for FL task acceptance and pricing, we limit our consideration to the terms involving these two decision variables on the right-hand side of Eq. \eqref{eq:goal}. Combining the expression for $f(r_i(t), p_i(t))$ from Eq. \eqref{eq:demand_for_service_1}, we have:
\begin{equation}
\small
\label{eq:goal_2_final}
\begin{aligned}
& \max \frac{1}{T} \sum_{t \in [0, T-1]}\sum_{i \in \mathcal{N}} \frac{x_i(t) \zeta_i p_i(t)}{(r_i(t))^{a_1}}[p_i(t) r_i(t) \rho_i(t) - q_i(t)],\\
& s.t. \quad \quad \quad \quad p_i(t) \geq p_i^{\min}, \forall t, \forall i,
\end{aligned}
\end{equation}
where $p_i^{\min}$ is the reserve price of data owner $i$. To get the optimal $p_i(t)$, we first get the first-order derivation of Eq. \eqref{eq:goal_2_final} with respect to $p_i(t)$ and then let the result equal to 0.
Then, we compute the optimal solution for $p_i(t)$ as: 
\begin{equation}
\small
\label{eq:price_optimal}
\begin{aligned}
p_i(t) = \max\left\{p_i^{\min}, \frac{q_i(t)}{2\rho_i(t)r_i(t)}\right\}.
\end{aligned}
\end{equation}
Eq. \eqref{eq:price_optimal} implies that DO $i$ should consider increasing its reserve price for new FL tasks when its current pending task queue is long, its current reputation is unfavorable, or its availability level is low (and conversely, reduce the reserve price under opposite conditions, while ensuring that its reserve price remains at least $p_i^{\min}$ to cover its costs). In scenarios where $i$ possesses a lower reputation, it is less likely to receive a substantial number of offers in AFL. Consequently, when others express interest in $i$'s services, it is in $i$'s best interest to charge a higher price to maximize the benefits it can derive from such opportunities.

To maximize Eq. \eqref{eq:goal_2_final}, the optimal value of $x_i(t)$ is:
\begin{equation}
\small
\label{eq:work_optimal}
\begin{aligned}
x_i(t) = 
 \begin{cases}
     1,  & \text{if $\rho_i(t)p_i(t)r_i(t) -  q_i(t) >  0$}, \\
     0, & \text{otherwise}.
 \end{cases}.
\end{aligned}
\end{equation}
\methodname{} ensures that the DO is never burdened with more FL tasks than it can manage at each time step. Therefore, when $x_i(t) = 1$, \methodname{} accepts up to $\theta_i^{\max}$ units of new FL task offerings into $i$’s pending task queue.

\section{Experimental Evaluation}
\subsection{Experiment Setup}
\textbf{Datasets.}
Our experiments are based on six commonly adopted datasets in FL: MNIST\footnote{http://yann. lecun. com/exdb/mnist/}, CIFAR-10\footnote{https://www.cs.toronto.edu/~kriz/cifar.html},
Fashion-MNIST (FMNIST) \cite{xiao2017fashion}, EMNIST-digits (EMNISTD) / letters  (EMNISTL) \cite{cohen2017emnist}, Kuzushiji-MNIST (KMNIST) \cite{clanuwat2018deep}. The FL models used are identical to those employed in \cite{tang2023utility}.
The AFL ecosystem under consideration comprises a total of 100 DOs and 6 MUs. To establish trust relationships among DOs, we leverage the Erdos-Renyi random graph with a connection probability of 0.7. The number of data samples owned by each DO is randomly generated within the range of 1,000 to 10,000. Each of the 6 MUs adopts a distinct bidding strategy, namely random, greedy, lin, bmub, fedbidder-simple, and fedbidder-complex \cite{tang2023utility}, respectively.

\textbf{Comparison Approaches.}
To the best of our knowledge, no existing decision support method has been designed for AFL DOs. As a result, we select common strategies, namely the random and greedy methods, as two of our baselines. 
Decision support for DOs predominantly revolves around pricing and sub-delegation choices. For the pricing decision, randomness is introduced (the prices for FL tasks follow a uniform distribution), while for the sub-delegation decision, both random and the greedy approaches are adopted (i.e., either to sub-delegate and the quantity of FL tasks to be sub-delegated are randomly determined, or to set the sub-delegation of FL tasks as extensively as possible). 
We then derive two experimental strategies: 1) Random pricing + random sub-delegation (\textbf{Rand-Rand}), 2) Random pricing + greedy sub-delegation (\textbf{Rand-Greedy}). Following \cite{tang2023utility}, we adapt the max utility bid method
for helping MUs bid for DOs by setting bid prices randomly within the range of 0 and the value of the DOs, and define a minimum pricing strategy (\textbf{Ampp}), i.e., the price is set to be randomly above the minimum price.
We also adapt the linear-form bid designed for MUs \cite{tang2023utility}, which sets the bid price directly proportional to the value of the DO (i.e., the reserve price), and formulate a linear-form pricing strategy (\textbf{Lin}) as setting the price to be directly proportional to the reserve price.
Both the \textbf{Ampp} and \textbf{Lin} pricing strategies can be combined with either random sub-delegation or greedy sub-delegation, resulting in another four strategies: \textbf{Ampp - Rand}, \textbf{Ampp - Greedy}, \textbf{Lin - Rand}, and \textbf{Lin - Greedy}.

\textbf{Evaluation Metrics.}
We employ two metrics for comparison among the alternative strategies: 1) MU Utility (\textbf{Utility}): This metric, defined in Eq. \eqref{eq:utility}, evaluates how much benefits DOs can derive under different strategies. It gauges system-level effectiveness from the viewpoint of pricing and sub-delegation. 2) Averaged MU FL model Test Accuracy (\textbf{Acc}): Any method integrated into the AFL ecosystem should foster sustainable development without undermining the interests of other stakeholders. Therefore, we adopt the test accuracy of resulting FL models obtained by MUs within the AFL ecosystem as a metric to evaluate the effectiveness of the DO decision support strategies.

\subsection{Results and Discussion}
Each experimental configuration was executed ten times, and the average results are shown in Table \ref{tab:performance}.
It can be observed that \methodname{} consistently performs better than all the baseline methods in terms of the utility gained by the DOs. In particular, \methodname{} improves the utility by 28.77\% on average compared with the best-performing baseline. The results demonstrate that \methodname{} provides effective decision support for DOs to help deal with the dynamic and challenging situations facing them in AFL. 
In general, approaches with the Ampp pricing strategy (Ampp-Rand and Ampp-Greedy) perform worse than those with other pricing strategies. In scenarios with Ampp pricing strategy, DOs tend to be more likely to set high asking prices in the auction market, resulting in fewer FL task requests from MUs and consequently, lower incomes for DOs. In addition, approaches based on greedy sub-delegation methods generally perform worse than those based on random sub-delegation methods in terms of utility. This is due to the fact that greedy sub-delegation tends to lead to higher costs for FL tasks, reducing the utility.
In addition, \methodname{} improves the test accuracy of the FL model achieved by 2.64\% on average. 

\textbf{Ablation Study}: 
We created two groups of ablated versions of \methodname{}:
    1) \textbf{w/o pricing}: In these ablated versions, we replace the pricing method with random pricing (\textbf{w/o pricing (R)}), above minimum pricing (\textbf{w/o pricing (A)}), and linear-form pricing (\textbf{w/o pricing (L)}), respectively.
    2) \textbf{w/o sub-delegation}: In these ablated versions, we replace the sub-delegation method with random sub-delegation (\textbf{w/o sub-delegation (R)}) and greedy sub-delegation (\textbf{w/o sub-delegation (G)}), respectively. 
The results are presented in Table \ref{tab:performance}. It can be observed that \methodname{} outperforms its ablated variants in terms of the total utility of both evaluation metrics. Therefore, the two designs by \methodname{} (i.e., the pricing strategy defined in Eq. \eqref{eq:price_optimal} and sub-delegation strategy defined in Eq. \eqref{eq:optimal_s}) key to its performance.

\section{Conclusions}
This paper frames the AFL from the perspective of DOs and investigates joint decision support for them by proposing \methodname{}. 
It leverages Lyapunov optimization to enable each DO to take on multiple FL tasks simultaneously to earn higher income for DOs and enhance the throughput of FL tasks in the AFL ecosystem.
The decisions are made based on a DO's current reputation, pending FL tasks, availability to engage in FL model training, and trust relationships with other DOs. It offers a systematic framework for DOs to make informed decisions regarding FL training task acceptance, sub-delegation, and pricing. \methodname{} is designed to maximize utility while building high-performance FL models. 

\section{Acknowledgments}
This research is supported, in part, by the National Research Foundation Singapore and DSO National Laboratories under the AI Singapore Programme (No. AISG2-RP-2020-019); the RIE 2020 Advanced Manufacturing and Engineering (AME) Programmatic Fund (No. A20G8b0102), Singapore; and the Natural Sciences and Engineering Research Council of Canada (NSERC).

\bibliographystyle{IEEEbib}
\bibliography{icme23}


\end{document}